\documentclass[journal]{IEEEtran}
\usepackage{tabularx}
\PassOptionsToPackage{table}{xcolor}
\usepackage{xcolor}
\usepackage{booktabs}
\usepackage{graphicx}

\definecolor{LightGray}{RGB}{230,230,230} 
\definecolor{SkyBlue}{RGB}{135,206,235} 
\definecolor{MintCream}{RGB}{220,255,220} 
\ifCLASSINFOpdf
\else
\fi
\begin{document}
%
\title{Harnessing Large Language Models' Empathetic Response Generation Capabilities for Online Mental Health Counselling Support}
%
%
%

\author{Siyuan Brandon Loh, and
        Aravind Sesagiri Raamkumar
\thanks{Siyuan Brandon Loh, and Aravind Sesagiri Raamkumar are with the Institute of High Performance Computing, Agency for Science, Technology and Research, 16-16 Connexis North, 1 Fusionopolis Way, Singapore
138632 }}

%
%

\markboth{Journal of \LaTeX\ Class Files,~Vol.~14, No.~8, August~2015}%
{Shell \MakeLowercase{\textit{et al.}}: Bare Demo of IEEEtran.cls for IEEE Journals}
%



\maketitle

\begin{abstract}
Large Language Models (LLMs) have demonstrated remarkable performance across various information-seeking and reasoning tasks. These computational systems drive state-of-the-art dialogue systems, such as ChatGPT and Bard. They also carry substantial promise in meeting the growing demands of mental health care, albeit relatively unexplored. As such, this study sought to examine LLMs' capability to generate empathetic responses in conversations that emulate those in a mental health counselling setting. We selected five LLMs: version 3.5 and version 4 of the Generative Pre-training (GPT), Vicuna FastChat-T5, Pathways Language Model (PaLM) version 2, and Falcon-7B-Instruct. Based on a simple instructional prompt, these models responded to utterances derived from the EmpatheticDialogues (ED) dataset. Using three empathy-related metrics, we compared their responses to those from traditional response generation dialogue systems, which were fine-tuned on the ED dataset, along with human-generated responses. Notably, we discovered that responses from the LLMs were remarkably more empathetic in most scenarios. We position our findings in light of catapulting advancements in creating empathetic conversational systems.
\end{abstract}


%
\IEEEpeerreviewmaketitle

\section{Introduction}
%
%
%
%

\IEEEPARstart{H}{umanity} faces an unprecedented need for mental health services. Global crises, such as the recent COVID-19 pandemic, have greatly burdened people's mental health, with the World Health Organization (WHO) reporting a 25\% increase in depression and anxiety cases during the first week of the pandemic. The accessibility of mental health services is far from ideal, with those at greatest risk of mental distress being the least likely to receive help \cite{world2022world}. This escalating demand for mental health services and workers highlights the urgent need for accessible, scalable, and transformative approaches to address the mental health crisis \cite{apa2021apa}. This demand is backed by the finding that mental health workers are more empathetic towards victims than general physicians and non-medical workers \cite{Santamaría-García2017}. Empathy is vital in these settings as it leads to higher satisfaction and improved patient outcomes \cite{YU2022}.

Digital technologies such as dialogue/conversational systems (i.e., chatbots) present viable solutions for providing remote psychological care and emotional support \cite{andersson2016internet}. Preliminary reports suggest positive outcomes for individuals who engage with such tools \cite{daley2020preliminary}. These automated solutions are also positively received by both general users and mental health professionals alike \cite{abd2021perceptions}\cite{sweeney2021can}. A recent study comparing physician and chatbot (ChatGPT) responses to patient questions in social media forums, found that chatbot responses had better quality and empathy \cite{10.1001/jamainternmed.2023.1838}. Apart from fully automated solutions, conversational AI systems have been found to be helpful in assisting novice counsellors in online peer support systems \cite{hsu2023helping}. Given their acceptance and positive results derived from digital platforms, it seems worthwhile to employ the latest advancements in artificial intelligence (AI) to enhance these initiatives further.

\subsection{Empathetic Conversational Systems}
Advancements in AI have paved the way for the development of dialogue systems imbued with the capacity to discern and appropriately respond to the emotional content of a user's messages. Termed Empathetic Conversational Systems (ECS)\cite{sesagiri2022empathetic}, these systems often represent a sophisticated modification of pre-trained encoder-decoder Transformer-based neural architectures \cite{vaswani2017attention}. Certain models include a dedicated function to encode the emotional content of a user's message \cite{kim2021perspective}, while others utilize external knowledge structures such as knowledge graphs to derive meaningful insights from a user's message that go beyond its immediate interpretation \cite{li2022kemp}. The emphasis of these systems on modelling empathetic responses, a crucial element in fostering therapeutic results in psychotherapy \cite{elliott2011empathy}, positions them as promising tools for technologically-mediated mental healthcare.

Despite their potential, the development of ECS is significantly constrained by the lack of high-quality training data. As pointed out by Raamkumar and Yang \cite{sesagiri2022empathetic}, the primary resource for developing ECS is the EmpatheticDialogues (ED) dataset \cite{rashkin2018towards}. This publicly available seminal dataset was designed to enable the development of dialogue agents capable of discerning feelings during a conversation and responding empathetically. However, the ED dataset presents several challenges.

The data in the ED dataset consists of conversations between randomly selected Mechanical Turk (mTurk) workers, without any criteria requiring participation from trained mental health professionals. This introduces a potential for significant variance in the types of responses in the dataset, increasing the risk of inclusion of malicious, judgemental, or unempathetic responses. Montiel and colleagues' findings support this concern \cite{montiel2022explainable}; Volunteers who scored high on an emotional quotient test rated the empathy level in a representative subset of responses in the ED dataset as significantly lower than those initially assigned in the dataset. Furthermore, the structure of the conversations within the ED dataset poses additional limitations. Most conversations in the dataset are brief, typically only encompassing one exchange, or 'turn'. This brevity leaves little room for an extended dialogue, which is a crucial component for modeling the different stages of dialogue typically encountered in counselling or mental health settings.  This could hinder the system's capability to fully engage with users and navigate the various stages of a therapeutic conversation.  Taken together, the variance in responses and the structure of the dataset underscores the shortcomings in ED. These limitations could result in ECS models that fall short of providing the needed empathetic responses and potentially negatively impact user engagement and trust in such systems.

\subsection{Large Language Models (LLMs)}

LLMs such as Generative Pre-Training models (GPT) \cite{brown2020language} have shown impressive capabilities across multiple tasks, including logical reasoning, text summarisation, machine translation, and language understanding \cite{kojima2022large}\cite{zhou2023lima}. GPT is the backbone of ChatGPT, the well-acclaimed general purpose chatbot.  Crucially, humans preferred responses from language models trained with minimal fine-tuning than those that were fine-tuned with human feedback \cite{zhou2023lima} [provide study context]. Overall, they showed that LLM's performance is highly dependent on the unsupervised, task-agnostic, pre-training phase, where the model encodes a general-purpose representation of a large quantity of text, rather than during the fine-tuning phase. This discovery, along with many others, suggests the potential for LLMs to serve as a practical alternative to ECSs in a mental health context, especially considering the data constraints discussed earlier.

Given the paucity of research in this domain, it remains to be seen if LLMs are capable of generating responses in a manner appropriate for a mental healthcare setting. Thus, the current study attempts to answer this central research question through a comparative evaluation of responses from ECS models and LLMs to a query in the ED dataset. The comparison is conducted at both the individual model level and the aggregated group level.  Each model's response was evaluated using a preexisting computational framework for detecting the presence of empathetic features in textual data\cite{sharma2020computational}. This framework, which models empathy in text as a three-dimensional construct, is used as a basis to answer our main research question (see Methods for details).

\section{Methods}
\subsection{Dataset}
We comparatively evaluated the empathetic response generation abilities of different language models through a series of experiments on the EmpatheticDialogues (ED) dataset (Rashkin et al., 2019). ED comprises a series of conversations between two participants. The first participant (P1) was randomly assigned one of 32 emotion words (the “prompt”) and was asked to recount a personal experience related to that emotion (the “situation”). The participant then entered a chatroom, where he/she discussed the “situation” with another participant (P2), who was tasked to listen and respond with empathy. Altogether, 810 individuals participated in the dataset creation exercise, amounting to 24,850 conversations. The dataset is split approximately into 80\% train, 10\% validation, and 10\% test partitions. We used the dialogues from the test partition for our experiments.

\subsection{Models}
\subsubsection{Large Language Models (LLMs)}
\begin{itemize}
    \item \textbf{Generative Pre-trained Transformer 3.5-Turbo (GPT-3.5)}: GPT-3.5 is a 345 billion parameter LLM trained on a large corpus of text on the internet\cite{brown2020language}.
    \item \textbf{Generative Pre-trained Transformer 4 (GPT-4)}: GPT-4 is the latest iteration of the GPT series from OpenAI. The intended improvements, scale, and the exact capabilities of GPT-4 are not yet fully disclosed due to its developmental stages.
    \item  \textbf{VicunaT5}: Vicuna FastChat-T5 is a chatbot trained unsupervised on 70,000 user-shared conversations \cite{zheng2023judging}.
    \item  \textbf{PaLM2}: Pathways Language Model (PaLM) 2 is a recent LLM developed by Google. We use the chat-bison-001 variant of the PaLM model since it has optimized for conversations \cite{anil2023palm}.
    \item \textbf{Falcon7I}: Falcon-7B-Instruct is a model based on the Falcon-7B LLM, finetuned with a mixture of chat and instruction datasets \cite{TechnologyInnovationInstitute}.

\end{itemize}

We prompted each LLM to elicit empathetic responses using the text prompt "This experiment requires you to continue the conversation with a user. The user is confiding in you on a personal matter. Listen with empathy. Avoid coming off as judgemental or apathetic".

\subsubsection{Empathetic Conversational Systems (ECS)}
\begin{itemize}
    \item \textbf{Knowledge Bridging for Empathetic Dialogue Generation (KEMP)}: KEMP is an external knowledge-driven empathetic dialogue system that uses information from knowledge graphs and emotion lexicons to encode the dialogue history. An attention-based decoder then generates the response, conditioned on the encoded content \cite{li2022kemp}.
    \item \textbf{Focused Empathy (FE)}:  Inspired by the Rational Speech Acts \cite{frank2012rsa}, FE is an empathetic dialogue system that reasons about the emotional state of its user before generating a response that is both conditioned on the perceived emotional state and the user's beliefs and perceptions of the response \cite{kim2021perspective}. 
    \item  \textbf{Cognitive Affective Behaviour (CAB)}:  CAB is a variant deep probabilistic generative model. It is made up of multiple modules, each designed to infer cognitive, behavioural, or affective information from a given piece of text \cite{gao2023cab}.  
\end{itemize}

We fine-tuned each ECS model on the ED training dataset using the code provided by the respective authors. 

\begin{table*}[h!]
\caption{\label{tab:eg} Example user prompt in ED where participant 1 was assigned with the emotion word "lonely", along with the corresponding human and model responses}
\centering
\begin{tabularx}{\textwidth}{|l|X|}
\hline
\textbf{Prompt} & My wife has been in Germany for 2 months.  I can't wait till she gets home. (emotional context: lonely)\\
\hline
\multicolumn{2}{|c|}{\textbf{Responses}} \\
\hline
\textbf{Human} &  Wow, i bet you miss her so much. My wife is going overseas with our son and they will be gone for 1 month! I know i will be miserable without them \\
\hline
\textbf{Retrieval} & When will she be returning ? \\
\hline
\textbf{KEMP} & that is awesome ! i hope you have a good time !\\
\hline
\textbf{FE} & germany is a beautiful place \\
\hline
\textbf{CAB} & that is awesome ! i am happy for you .\\
\hline  
\textbf{GPT3.5} &  I understand how you feel. Long-distance relationships can be tough and I'm sure you miss her a lot. How have you been coping with her absence?" \\
\hline
\textbf{GPT4} & That sounds like a long time to be away from someone you love. It must be challenging at times. How have you been keeping in touch in her absence? Do you plan to celebrate in any special way when she gets home?\\
\hline
\textbf{VicunaT5} & I understand how you feel. I'm here for you and I'm here to listen and offer any support I can. How are you doing these days and how long has she been away?\\
\hline
\textbf{Falcon7I} & I'm sorry to hear that. It's important to communicate with your wife about how you're feeling. I'm here to listen if you need someone to talk to.\\
\hline
\textbf{PaLM2}  &  I can imagine how difficult it must be to be apart from your wife for so long. I'm sure you're missing her a lot. It's great that you're looking forward to her return. I hope the time goes by quickly for you.\\
\hline
\end{tabularx}
\end{table*}

\begin{table*}[h!]
\caption{\label{tab:sentiment} Prompts grouped by sentiment}
\centering
\begin{tabularx}{\textwidth}{|l|X|}
\hline
\textbf{Sentiment} & \textbf{Prompt} \\
\hline
Positive (n=1040) & nostalgic, caring, sentimental, confident, content, excited, joyful, impressed, proud, faithful, grateful, hopeful \\
\hline
Negative (n=1157)& afraid, anxious, terrified, angry, disgusted, annoyed, furious, ashamed, guilty, devastated, disappointed, lonely, sad, embarrassed \\
\hline
Ambiguous (n=348)& surprised, apprehensive, anticipating, prepared \\
\hline
\end{tabularx}
\end{table*}

\subsubsection{Baselines}
\begin{itemize}
    \item \textbf{Human} : Original human responses from the ED dataset \cite{rashkin2018towards}. Even though these are actual human responses, we will refer to this baseline as a 'human' model for the sake of reference.
    \item \textbf{ED-Retrieval}: The baseline model published in the ED dataset paper \cite{rashkin2018towards}. In this model, transformer-based networks encode the dialogue history and a set of candidate responses. The candidate whose encoded state has the greatest dot product with the dialogue history is subsequently chosen as the model’s response. Similar to ECSs, ED-Retrieval was fine-tuned on the ED training set using the code provided.
\end{itemize}

\subsection{Experimental Setup}
Each model responded to the first utterance of each conversation in ED's test dataset (\textit{n} = 2,545). A sample scenario from the ED dataset with model responses is provided in Table \ref{tab:eg}. Responses were subsequently evaluated using three metrics that were designed to measure the empathetic ability of counsellors in online forums \cite{sharma2020computational}. The first metric codes for the presence of linguistic markers indicative of a help-seekers' attempt to address the emotional concerns of the person in distress (\textbf{`Emotional Reactions'}). The second metric codes for linguistic markers suggestive of a help-seeker's attempt to restate the presenting problems of the person in distress (\textbf{`Interpretations'}). The final metric codes for linguistic markers that highlight the help-seeker's attempt to dive deeper into topics that the person in distress presents (\textbf{`Exploration'}). These metrics take on three discrete labels that denote the strength of the respective signal in a given piece of text (none, weak, strong). Three GPT3 models were fine-tuned on the original dataset provided by the original authors to classify text with respect to each metric \cite{sharma2020computational} (see Table \ref{tab:sentiment} for positive/negative responses from the original dataset).

Since our primary interest lay not in discerning the degrees of 'weak' and 'strong', but rather in determining the presence or absence of the outcome, we consolidated the 'weak' and 'strong' groups into a single unified category. This effectively transformed our dependent variable into a binary logistic format, allowing us to focus on the critical distinction - the absence ('none') vs. the presence ('weak' or 'strong') of a particular empathy metric across different groups.

\subsection{Statistical Analysis}
 We examined group-level differences across conversational contexts by grouping each conversation based on the "prompt" that was assigned to participant P1. Notably, we categorized each "prompt" as conveying either positive, negative, or ambiguous sentiment (Table \ref{tab:sentiment}). This new sentiment variable enables us to observe how responses from model types differ across the sentiment undertones of the conversation. Separate logistic mixed models were fitted to each empathy metric, using model type, sentiment, and their interactions as predictors. We also included a random intercept for each model to account for their idiosyncratic effects on the scores for each empathy dimension.

\section{Results}

\subsection{Descriptive Statistics}
The following section characterizes the nature of responses by each model, according to each respective empathy metric (Table \ref{table:descriptives}). For the Emotional Reaction metric,  VicunaT5  from the LLM group has the greatest proportion of responses perceived to contain features aimed at catering to the emotions of the other interlocutor (0.682), followed by PaLM2 (0.585) and Falcon7I (0.548).

\begin{table}[h!]
\centering
\caption{Models' responses rated across empathy metrics. Score represents the proportion of responses (n=2545) containing empathetic features. blue = baseline, grey= ECS, green = LLM}
\label{table:descriptives}
\begin{tabular}{|c|c|c|c|c|}
\hline
\textbf{Model} & \textbf{Emotional Reaction} & \textbf{Interpretation} & \textbf{Exploration} \\
\hline
\rowcolor{SkyBlue}ED & 0.249 & 0.419 & 0.278 \\
\hline
\rowcolor{SkyBlue}Human & 0.140 & 0.596 & 0.247 \\
\hline
\rowcolor{LightGray}CAB & 0.281 & 0.397 & 0.345 \\
\hline
\rowcolor{LightGray}FE & 0.240 & 0.400 & 0.299 \\
\hline
\rowcolor{LightGray}KEMP & 0.334 & 0.464 & 0.263 \\
\hline
\rowcolor{MintCream}Falcon7I & 0.548 & 0.268 & 0.417 \\
\hline
\rowcolor{MintCream}GPT3.5 & 0.339 & 0.623 & 0.205 \\
\hline
\rowcolor{MintCream}GPT4 & 0.411 & 0.822 & 0.267 \\
\hline
\rowcolor{MintCream}PaLM2 & 0.585 & 0.217 & 0.532 \\
\hline
\rowcolor{MintCream}VicunaT5 & 0.682 & 0.609 & 0.369 \\
\hline
\end{tabular}
\end{table}

\begin{figure*}[ht]
\caption{Average proportion of responses in each model type with empathetic features. Scores are grouped by sentiment. Top panel: Emotional Reaction. Middle panel: interpretation. Bottom panel: Exploration}
\label{fig:sentimodeltype}
\centering
\includegraphics[width=\linewidth]{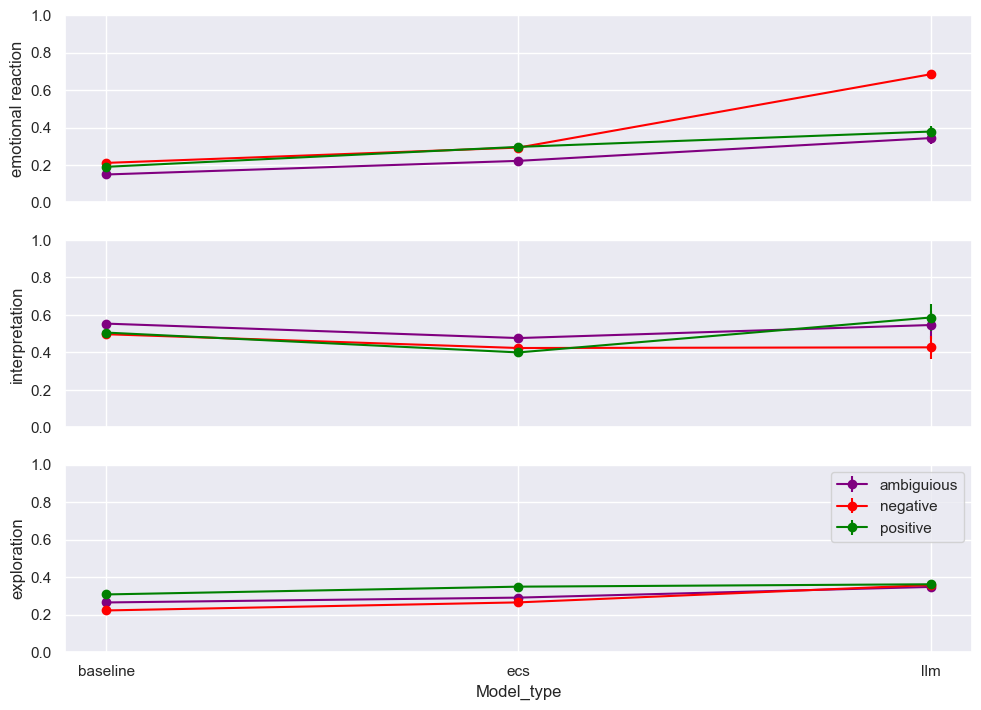}
\end{figure*}

\begin{table*}[htbp]
\centering
\caption{Results for Logistic Mixed Effects Model of Emotional Reaction}
\label{tab:ER}

\begin{tabular}{@{}llll@{}}
\toprule
Predictors & Odds Ratios & CI & p \\
\midrule
(Intercept) & 0.17 & 0.09 -- 0.32 & $<0.001$ \\
model type (MT) [ecs] & 1.68 & 0.73 -- 3.90 & 0.225 \\
model type [llm] & 2.96 & 1.37 -- 6.40 & 0.006 \\
sentiment (Sent) [negative] & 1.54 & 1.22 -- 1.94 & $<0.001$ \\
sentiment [positive] & 1.35 & 1.06 -- 1.70 & 0.014 \\
MT [ecs] $\times$ Sent [negative] & 0.95 & 0.71 -- 1.26 & 0.703 \\
MT [llm] $\times$ Sent [negative] & 3.01 & 2.32 -- 3.90 & $<0.001$ \\
MT [ecs] $\times$ Sent [positive] & 1.10 & 0.83 -- 1.47 & 0.513 \\
MT [llm] $\times$ Sent [positive] & 0.87 & 0.67 -- 1.14 & 0.319 \\
\midrule
Random Effects & & & \\
ICC & 0.06 & & \\
Observations & 25357 & & \\
Marginal R$^2$ / Conditional R$^2$ & 0.158 / 0.206 & & \\
\bottomrule
\end{tabular} 
\end{table*}

\begin{table*}[ht]
\centering
\begin{minipage}[b]{.5\linewidth}
\centering
\caption{Results for Logistic Mixed Effects Model of Interpretation}
\label{tab:IP}
\begin{tabular}{@{}lllll@{}}
\toprule
Predictors & Odds Ratios & CI & p \\
\midrule
(Intercept) & 1.25 & 0.43 -- 3.59 & 0.683 \\
model type (MT) [ecs] & 0.73 & 0.19 -- 2.86 & 0.650 \\
model type [llm] & 1.01 & 0.29 -- 3.56 & 0.987 \\
sentiment (Sent) [negative] & 0.79 & 0.67 -- 0.94 & 0.008 \\
sentiment [positive] & 0.82 & 0.69 -- 0.98 & 0.026 \\
MT [ecs] $\times$ Sent [negative] & 1.02 & 0.82 -- 1.27 & 0.868 \\
MT [llm] $\times$ Sent [negative] & 0.69 & 0.56 -- 0.85 & 0.001 \\
MT [ecs] $\times$ Sent [positive] & 0.89 & 0.71 -- 1.12 & 0.316 \\
MT [llm] $\times$ Sent [positive] & 1.50 & 1.21 -- 1.86 & $<0.001$ \\
\midrule
Random Effects \\
ICC & 0.16 & & \\
Observations & 25357 & & \\
Marginal R$^2$ / Conditional R$^2$ & 0.027 / 0.178 & & \\
\bottomrule
\end{tabular}
\end{minipage}%
\begin{minipage}[b]{.5\linewidth}
\centering
\caption{Results for Logistic Mixed Effects Model of Exploration}
\label{tab:EX}
\begin{tabular}{@{}llll@{}}
\toprule
Predictors & Odds Ratios & CI & p \\
\midrule
(Intercept) & 0.36 & 0.21 -- 0.61 & $<0.001$ \\
model type (MT) [ecs] & 1.13 & 0.56 -- 2.28 & 0.725 \\
model type [llm] & 1.43 & 0.75 -- 2.69 & 0.275 \\
sentiment (Sent) [negative] & 0.79 & 0.65 -- 0.96 & 0.019 \\
sentiment [positive] & 1.23 & 1.02 -- 1.50 & 0.032 \\
MT [ecs] $\times$ sent [negative] & 1.11 & 0.87 -- 1.42 & 0.406 \\
MT [llm] $\times$ sent [negative] & 1.32 & 1.06 -- 1.66 & 0.015 \\
MT [ecs] $\times$ sent [positive] & 1.06 & 0.83 -- 1.36 & 0.628 \\
MT [llm] $\times$ sent [positive] & 0.86 & 0.69 -- 1.08 & 0.205 \\
\midrule
Random Effects & & & \\
ICC & 0.04 & & \\
Observations & 25357 & & \\
Marginal R$^2$ / Conditional R$^2$ & 0.014 / 0.054 & & \\
\bottomrule
\end{tabular}
\end{minipage}
\end{table*}

The scores for the other LLMs were not much behind. Most notably, all LLMs outperformed the remaining models in the other two groups. On the other hand, the 'Human' model from the baseline group has the lowest score of 0.140, signifying less effective emotional resonance.

For the Interpretation metric, GPT-4 (0.822) performed the best, while PaLM2 (0.217) performance was the worst amongst all models from three groups. The second and third best performance is from GPT3.5 (0.623) and VicunaT5 (0.609). Both PaLM2 and Falcon7I (0.268) had the lowest scores. These findings suggest substantial within-group variation within the LLM models group in the tendency for inquiry and reinterpretation of the other interlocutor's concerns. In the baseline group, the human baseline (0.596) was better than all the models in the ECS group.

Models from the LLM group, namely PaLM2 (0.532) and GPT-3.5 (0.205) were respectively the highest and lowest scoring models on the Exploration metric. The second best LLM is Falcon7I (0.417). Different from the emotional reaction metric, there is a visible variance in the scores of the different LLMs. In fact, the two GPT models have the lowest scores across all the models in the three groups.  Apart for these two LLMs, the other LLMs have higher scores than the models in the other two groups.

\subsection{Do models from different groups differ in their empathetic response capabilities?}
We report the results of our statistical model of each empathy metric in the following sections. Tables \ref{tab:ER}, \ref{tab:EX}, \ref{tab:IP} display the odds ratio for each predictor of the mixed effects model. In the context of the current study, the odds ratio compares the odds that a response by a model in a particular group obtains a positive score on any given metric to those from the baseline model. An odds ratio higher than 1 denotes that the model type of interest is more likely than the baseline model to obtain a positive score on the metric, while an odds ratio lower than 1 implies otherwise.

Overall, marginal $R^2$ for all linear mixed effect models are .206, .054, and .178 for Emotional Reaction, Interpretation, and Exploration metrics, respectively. This finding indicates that at best, 20.6 \% of score on each metric can be explained by sentiment and model type, along with their interaction.

Intraclass correlations, which measure the ratio of between-group variance to the total variance are .06, .04, and .16 for Emotional Reaction, Interpretation, and Exploration metrics, respectively. This finding indicates a substantial amount of within-group variance in models' performance across each metric. 

\subsubsection{Emotional Reaction (Table \ref{tab:ER})}
Results from the linear mixed effect model indicated that LLMs generated significantly more responses that catered to the user's emotions than the baseline models (odds ratio: 2.96 [1.37 - 6.4], p < .005). There was no statistically significant difference between the responses from ECSs and baseline models (odds ratio: 1.68 [.73 - 3.9], p = .225).

The coefficients for interaction between model type and sentiment revealed interesting trends. Although most factors were statistically insignificant, we observe that the interaction between LLMs and prompts in the negative sentiment condition has a significant effect on the Emotional Reaction metric (odds ratio: 3.01 [2.32 - 3.90], p < .001). This finding suggests that LLMs' responses were more likely to be rated as containing Emotional Reactions when they replied to negative sentiment messages than the baseline models. 

\subsubsection{Interpretation (Table \ref{tab:IP})}
Results from the linear mixed effect model indicated that model type did not significantly influence scores on the Interpretation metric. The odds ratio for both ECS (odds ratio: .73 [.19 - 2.86], p = .65) and LLM (odds ratio: .1.01 [.29 - 3.56], p = .987) model types did not statistically differ from 1.

Coefficients for interaction between model type and sentiment revealed that the interaction between LLMs and prompts in both the negative (odds ratio: .79 [.67 - .94], p < .05) and positive (odds ratio: .82 [.69 - .98], p < .05) sentiment condition has a significant effect on the Interpretation metric. This finding suggests that LLMs were significantly less likely than the baseline models to interpret the meaning behind a message when it conveyed a negative sentiment, with the opposite being true for messages that conveyed a positive sentiment.

\subsubsection{Exploration (Table \ref{tab:EX})}
Results from the linear mixed effect model indicated that model type did not significantly influence scores on the Exploration metric. The odds ratio for both ECS (odds ratio: 1.13 [0.56 – 2.28], p = 0.725) and LLM (odds ratio: .1.01 [.29 - 3.56], p = .987) model types did not statistically differ from 1.

Coefficients for interaction between model type and sentiment were non-significant, with the exception of the interaction between LLMs and prompts in the negative (odds ratio: 1.32 [1.06 – 1.66] p < 0.05). This finding suggests that LLMs were significantly more likely than the baseline models to explore topics beyond the content of the immediate post when it conveyed a negative sentiment.

 \section{Discussion}
This study conducted a comprehensive evaluation of several automated language generation models including LLMs and traditional response generation conversational models, focusing on their ability to elicit empathetic responses. Overall, we found partial albiet promising support for our hypothesis; LLMs were significantly better at producing responses that signaled an attempt at catering to the feelings expressed by the user in their prompts than ECS models or our human-level baselines.  

On the Interpretation metric, LLMs produced better responses for positive emotion classes than for negative ones. This result is worth highlighting, given the prominence of negative emotions in mental health scenarios. Surprisingly, this is the only metric where the performance of the baseline group, which comprised human responses, was comparable to those from the models in the LLM and ECS groups. 

The human baseline, which comprised original responses from ED, demonstrated the worst performance for Emotional Reaction and Exploration metrics. This reflects our initial position concerning dataset quality and the downstream consequence it has on developing empathetic AI agents. Nonetheless, our findings offer evidence for the viability of a less "data-hungry" approach in light of the current dataset limitations \cite{zhou2023lima}. Here, the key takeaway is that pre-trained LLMs already possess a nuanced text representation that can be easily adapted to most downstream tasks. 

From these results, LLMs, as a result of their exposure to wide-ranging and complex training data, might be better poised for application in mental health care settings where adaptability, nuanced understanding, and empathetic response generation are paramount. Conversely, ECS, while displaying a balanced performance, do not outperform LLMs, possibly due to limitations or specificities in the scope of their training data. We believe that ECS models will be replaced by LLMs in the near future since LLMs are able to produce decent results with just simple prompts. There are other activities such as prompt engineering and fine-tuning which can further improve the performance of LLMs.

Our study's LLM results differ from the existing studies that demonstrated the superiority of GPT4 and GPT3.5 against other commercial and open-source LLMs for a wide range of tasks \cite{zheng2023judging}. We opine that GPT LLMs can potentially produce better results with differently framed prompts and different sets of evaluation metrics. Nonetheless, these findings imply a potential variance in the capability of LLMs, despite them being trained with similar methods and expansive data sets. LLM's performance could be further improved with more detailed prompts and finetuning on relevant datasets. 

Although the current analysis favors LLMs for potential application in mental health settings, it is imperative to acknowledge that the real-world implementation might carve out a different trajectory dictated by actual patient interactions and personalized responses. Secondly, the evaluation metrics used in this study are limited by the training dataset used for the three corresponding classifier models. A user-based evaluation could bring forth vastly different results. Nevertheless, clinicians and mental health workers will be able to embed personalized data and influence the responses generated by the LLMs to a great extent by adopting tools and systems which are based on retrieval-augmentend generation (RAG) \cite{lewis2020retrieval}, a method for using LLMs on top of local data.

\section{Conclusion}
Our analysis provides a preliminary basis for understanding the performance of LLMs as against traditional response generation models and human baselines within empathy-driven contexts. These insights underscore the importance of dataset diversity and interpretative sensitivity for AI models to optimally function within mental health care settings, thus providing an avenue for targeting future improvements in AI conversational models. In our future studies, we intend to further research in two directions. In the first direction, we will evaluate the performance of LLMs as assistive agents in helping counsellors who moderate online mental health help forums. In the second direction, we will include more open-source LLMs in the analysis and plan experiments leveraging different types of prompts, and fine-tuning approaches in order to attain more improvements in the LLMs performance. 

\bibliographystyle{IEEEtran}
\bibliography{references}

\section*{Acknowledgements}

This research is supported by the Agency for Science, Technology and Research (A*STAR) under its SERC Council Strategic Fund (C210415006).   The authors are grateful for the helpful discussion with Yinping Yang, and Raj Kumar Gupta. 

\section*{Author contributions statement}
All authors conceived and conducted the experiments, analysed the results, and reviewed the manuscript. 

\section*{Additional information}

To include, in this order: \textbf{Accession codes} (where applicable); \textbf{Competing interests} (mandatory statement).

\ifCLASSOPTIONcaptionsoff
  \newpage
\fi



%

%

\begin{IEEEbiographynophoto}{Siyuan Brandon Loh}
received his BA degree in psychology from Nanyang Technological University, Singapore. He is currently a research engineer with the Institute of High Performance Computing, A*STAR, Singapore. His research interest is in the application and development of robust computational systems to the study of social dynamics and behaviour. 

\end{IEEEbiographynophoto}


\begin{IEEEbiographynophoto}{Aravind Sesagiri Raamkumar}
 received the PhD degree in information studies from Nanyang Technological University, Singapore. He is currently a senior scientist with the Institute of High Performance Computing, A*STAR, Singapore. His research interests include conversational systems,  text mining, recommender systems, information retrieval, health informatics, social network analysis, scholarly metrics, deep learning.
\end{IEEEbiographynophoto}




\end{document}